\documentclass[runningheads]{llncs}
\usepackage[nottoc]{tocbibind}
\usepackage[english]{babel}
\usepackage{xcolor}
\usepackage{graphicx}
\usepackage[ruled,linesnumbered]{algorithm2e}
\usepackage{subfigure}
\usepackage{amsmath}
\usepackage{hyperref}
\usepackage{cite} 
\usepackage{float} %提供float浮动环境
\usepackage{booktabs} %提供命令\toprule、\midrule、\bottomrule
\begin{document}
\title{Satisfactory Medical Consultation based on Terminology-Enhanced Information Retrieval and Emotional In-Context Learning\thanks{These authors contributed to the work equally and should be regarded as co-first authors.\\ ** Corresponding author.}}
%
%\titlerunning{Abbreviated paper title}
% If the paper title is too long for the running head, you can set
% an abbreviated paper title here
%
\author{Jing Tang\inst{1,*} \and 
        Kaiwen Zuo\inst{2,*} \and 
        Hanbing Qin\inst{3} \and 
        Binli Luo\inst{1} \and
        Ligang He\inst{2}\and
        Shiyan Tang\inst{4,**}}

% Authorrunning is used for abbreviated author list in the header
\authorrunning{J. Tang et al.}

\institute{School of Mechanical Science and Engineering, Huazhong University of Science and Technology, Wuhan 430074, Hubei, China, \email{focusers@163.com} \and 
           School of Computer Science, University of Warwick, Coventry, CV4 7AL, UK, \email{Kaiwen.Zuo@warwick.ac.uk,ligang.he@warwick.ac.uk} \and 
           School of Software and Microelectronics, Peking University, Beijing 100871, China, \email{qinhanbing@stu.pku.edu.cn} \and 
           MRC Weatherall Institute of Molecular Medicine, University of Oxford, Oxford, OX1 2JD, UK, \email{shiyan.tang@ndcls.ox.ac.uk}}

\maketitle              % typeset the header of the contribution
\vspace{-10mm}
\begin{abstract}
Recent advancements in Large Language Models (LLMs) have marked significant progress in understanding and responding to medical inquiries. However, their performance still falls short of the standards set by professional consultations. This paper introduces a novel framework for medical consultation, comprising two main modules: Terminology-Enhanced Information Retrieval (TEIR) and Emotional In-Context Learning (EICL). TEIR ensures implicit reasoning through the utilization of inductive knowledge and key terminology retrieval, overcoming the limitations of restricted domain knowledge in public databases. Additionally, this module features capabilities for processing long context. The EICL module aids in generating sentences with high attribute relevance by memorizing semantic and attribute information from unlabelled corpora and applying controlled retrieval for the required information. Furthermore, a dataset comprising 803,564 consultation records was compiled in China, significantly enhancing the model's capability for complex dialogues and proactive inquiry initiation. Comprehensive experiments demonstrate the proposed method's effectiveness in extending the context window length of existing LLMs. The experimental outcomes and extensive data validate the framework's superiority over five baseline models in terms of BLEU and ROUGE performance metrics, with substantial leads in certain capabilities. Notably, ablation studies confirm the significance of the TEIR and EICL components. In addition, our new framework has the potential to significantly improve patient satisfaction in real clinical consulting situations.


\end{abstract}

\section{Introduction}
\vspace{-5mm}
Medical consultation has always been a cornerstone of healthcare services, providing necessary guidance and information to individuals seeking medical attention\cite{schwartz2023black}. Traditionally, these consultations occurred face-to-face within clinical settings. However, the uneven distribution of medical resources and an underdeveloped referral system in China have led to prolonged waiting times and limited consultation periods for many patients\cite{wang2023time}. Additionally, patients often encounter fear and anxiety when seeking medical help. Consequently, the language employed in communication during medical consultations holds great significance for both healthcare providers and patients\cite{wang2023time}. Should doctors fall short of meeting patients’ unrealistically high \
expectations during medical consultations\cite{schwartz2023black}, it can result in dissatisfaction among patients. Therefore, patients' emotions and satisfaction during these consultations should never be neglected. 

\begin{figure}[htbp]
\centering
% Increase the scale to make the image larger. Adjust the scale factor as needed.
\includegraphics[scale=0.55]{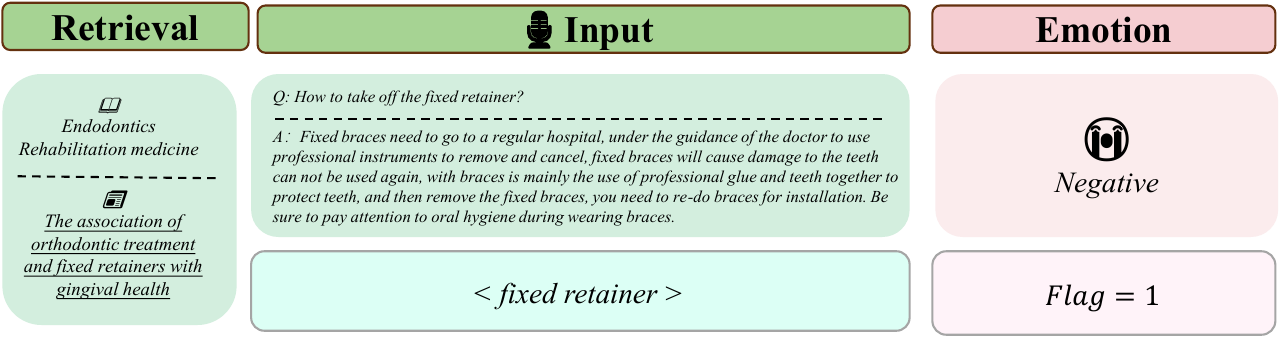}
% Alternatively, you can set a specific width or height.
% \includegraphics[width=1.5\linewidth]{QA table.pdf}
% \includegraphics[height=0.5\textheight]{QA table.pdf}
\caption{An example of model responses, which involves retrieving information about dental treatments, processing inquiries about these treatments, and identifying the emotional tone of the information presented.}
\label{fig:qatable}
\end{figure}

Nowadays, the rapid evolution of digital technology has ushered in a new era of medical service delivery. A significant proportion of medical consultations transpire online\cite{younis2024systematic,engelseth2021systems,alberts2023large}, offering unprecedented accessibility and convenience to patients. This digital transformation has not only redefined the interaction between healthcare providers and patients but has also generated a massive influx of data\cite{el2023integration}. This article delves into the realm of medical consultation in China by meticulously analyzing 803,564 instances of such data, collected through advanced web crawlers designed to navigate and retrieve information from a myriad of online consultation platforms. An example of medical dialogue, as illustrated in Figure~\ref{fig:qatable}, emphasizes that by engaging in proactive keywords and emotional inquiries, a more nuanced understanding of the patient's experience and potential concerns can be obtained, thereby contributing to a more comprehensive medical diagnosis.

The motivation for embarking on this research journey is rooted in the need to enhance the existing knowledge-enhanced information retrieval models prevalent in the field. While these models have significantly contributed to the advancement of medical consultations, they are not devoid of limitations. One of the primary concerns is the accuracy and relevance of the information provided during the consultations\cite{senbekov2020recent}. Misinformation or irrelevant advice can lead to patient dissatisfaction, mistrust, and in severe cases, adverse health outcomes\cite{wang2023ethical}. Additionally, the impersonal nature of online consultations often overlooks the emotional and psychological nuances of patient-provider interactions, further complicating the effectiveness of digital medical advisories.

Another pertinent issue is the efficiency of information retrieval during consultations. Patients seeking medical advice are often met with overwhelming amounts of information\cite{kenei2022supporting}, much of which may be irrelevant or too complex to understand. This not only prolongs the consultation process but also adds to the cognitive load of the patients, potentially exacerbating their stress and anxiety\cite{ho2022explainability}. The existing systems need to be more adept at quickly providing precise, understandable\cite{van2020online}, and relevant information tailored to the individual needs of each patient.

In light of these challenges, this article introduces a novel framework aimed at significantly enhancing the quality and effectiveness of online medical consultations. The proposed framework is the culmination of extensive research and development, leveraging the latest advancements in information technology, data analytics, and LLMs. The contributions of this study are manifold, addressing the core issues identified in the current systems and setting new benchmarks for what digital medical consultations can achieve. The enhancements are specifically designed to improve the accuracy, efficiency, and user satisfaction of online medical consultations, making healthcare advice more accessible and reliable for everyone. Our contributions are detailed as follows:

\begin{enumerate}
    \item\textbf{Generating Enhancement Based on Terminology-Enhanced Information Retrieval (TEIR)}: Traditional terminology match methods are often limited by their literal approach to query understanding, frequently overlooking the contextual nuances and user intent. Our approach revolutionizes this mechanism by incorporating advanced linguistic models and terminology-aware algorithms. These enhancements enable the system to obtain the underlying intent and specifics of user queries, leading to a more nuanced and accurate retrieval of information. This not only improves the relevance of the consultation content but also reduces the time users spend sifting through unnecessary information.
    \vspace{1em}
    \item \textbf{Generate Enhancements Based on Emotional In-Context Learning (ECIL)}: ECIL ensures that the information provided is coherent, logically structured, and easy to navigate. At the same time, emotional annotation adds a layer of emotional intelligence to the system, enabling it to recognize and respond to the emotional cues of the users. This not only makes the consultation more personalized and empathetic but also significantly enhances user engagement and satisfaction.
    
\end{enumerate}

% In the subsequent sections, the article will provide an in-depth exploration of each contribution, detailing the methodologies employed, the experimental setup, and the comprehensive analysis of the results. By dissecting the intricacies of the proposed enhancements and their impact on medical consultation, the study aims to contribute valuable insights and tools to the fields of medical informatics, digital health, and patient care. The ultimate goal is to pave the way for a future where digital medical consultations are as reliable, personalized, and satisfying as their traditional counterparts, thereby improving the overall quality of healthcare services and patient outcomes across the globe.
\vspace{-7mm}
\section{Related Work}
\vspace{-2mm}
\subsection{Keyword Retrieval Generation}
\vspace{-2mm}
Recently, many studies have provided a diverse look into the application of keyword retrieval generation in online medical consultations\cite{Ren2023RetrieveandSampleDE}. These cover areas from the training of large language models (LLMs) for medical chats to the systematic extraction of information from electronic medical records and the use of metadata and knowledge graphs for improved dialogue generation\cite{jiang2023think}. These insights are crucial for developing a new framework aimed at improving the accuracy, efficiency, and patient satisfaction of online medical consultations.

In the realm of enhancing online medical consultation, recent studies have introduced various innovative approaches. ChatDoctor\cite{li2023chatdoctor} details the training of a generic conversation model tailored for medical consultations, emphasizing advanced keyword retrieval techniques. Concurrently, Yu et al.\cite{yu2023leveraging} propose a taxonomy for the utility of ChatGPT in healthcare, highlighting the role of Large Language Models in keyword retrieval and information generation. Further, Ford et al. \cite{ford2016extracting} examine the use of US hospital-based EMRs, employing keyword searches and rule-based algorithms to enhance medical document retrieval, and Malet et al.\cite{malet1999model} discuss the use of Medical Core Metadata to improve keyword-based retrieval in medical consultations. IAG\cite{zhang2023iag} combines the advantages of retrieval and prompt methods, generates knowledge through keywords inductive enhancement, and inputs it into the generator along with the retrieved documents, leading to strong performance in open-domain question-answering tasks. Lastly, Nature introduces a method for knowledge graph enhanced medical dialogue generation, leveraging the Unified Medical Language System for more accurate and relevant medical dialogues\cite{varshney2023knowledge}.

Our proposed framework encompasses a multifaceted approach to enhance online medical consultations. It includes an advanced keyword retrieval generation system designed to improve the accuracy and relevance of information retrieval. Additionally, the framework integrates Template Engineering and Sentiment Annotation Generation, a feature that analyzes and responds to the emotional tone of patient inquiries, ensuring a more empathetic and personalized consultation experience. Further details of each component will be introduced and explored in the following sections.
\vspace{-2mm}
\subsection{Template Engineering and Sentiment Annotation Generation} 
\vspace{-2mm}
In the quickly emerging field of online medical consultation, the integration of various methods such as Template Engineering and Sentiment Annotation shows immense scope for the enhancement of accuracy, efficiency, and patient satisfaction\cite{vanAtteveldt2021TheVO, Yang2021ReExaminingTI}. This approach, as the proposed paper plans to further explore, primarily employs techniques previously explored in different fields, the most notable among them being sentiment analysis and multidimensional trust.

Wouter van Atteveldt et al. demonstrated the effectiveness of manual annotation and crowd coding in sentiment analysis\cite{vanAtteveldt2021TheVO}, a component that plays a pivotal role in gaining insights from patients' sentiments during online medical consultation. This historical emphasis on human-embedded sentiment analysis tools might be supplemented by the AraSenCorpus model introduced by Al-laith et al., which has redefined the scale of sentiment analysis with proficient sentiment annotation on a large-scale text corpus\cite{Allaith2021AraSenCorpusAS}. GRACE memorizes semantic and attribute information from unmarked corpora and obtains the required information through controllable re-retrieval\cite {inproceedings}.

Online medical consultation heavily relies on a patient's trust, a significantly multidimensional element\cite{Yang2021ReExaminingTI}. Trust bears an impact on patients' decision to continue the use of online services, and as shown by Yang et al., the incorporation of both technological and interpersonal aspects improves the chances of service continuity\cite{Yang2021ReExaminingTI}. Further, analysis by Wan et al. revealed other influencing factors such as doctors' soft skills and overall service experience that enhance the volume of doctor consultation on online platforms\cite{Wan2021InfluencingFA}.

In addition to trust, efficiency and patient satisfaction have been recognized as substantial factors contributing to the success of an online medical consultation framework. Using the employment of scribes as an example, Gidwani et al. demonstrated their positive impact on physician satisfaction and charting efficiency, without impinging upon patient satisfaction \cite{Gidwani2017ImpactOS}. The shifting dynamics were further reflected in Jiang et al.'s study, pointing out service-related features like service delivery quality and patient involvement to be more significant than physician reputation, in patients' payment decisions\cite{Jiang2020AnalysisOM}.

Overall, combining these aspects, the proposed method is well-supported by earlier research, and it is thoughtfully aimed at bringing a comforting user experience by improving the efficiency of medical consultation\cite{vanAtteveldt2021TheVO, Yang2021ReExaminingTI, Gidwani2017ImpactOS, Wan2021InfluencingFA, Allaith2021AraSenCorpusAS, Jiang2020AnalysisOM}.

\vspace{-4mm}
\section{Methodology}
\begin{figure}[htbp]\label{fig1}
\centering
% Increase the scale to make the image larger. Adjust the scale factor as needed.
\includegraphics[scale=0.55]{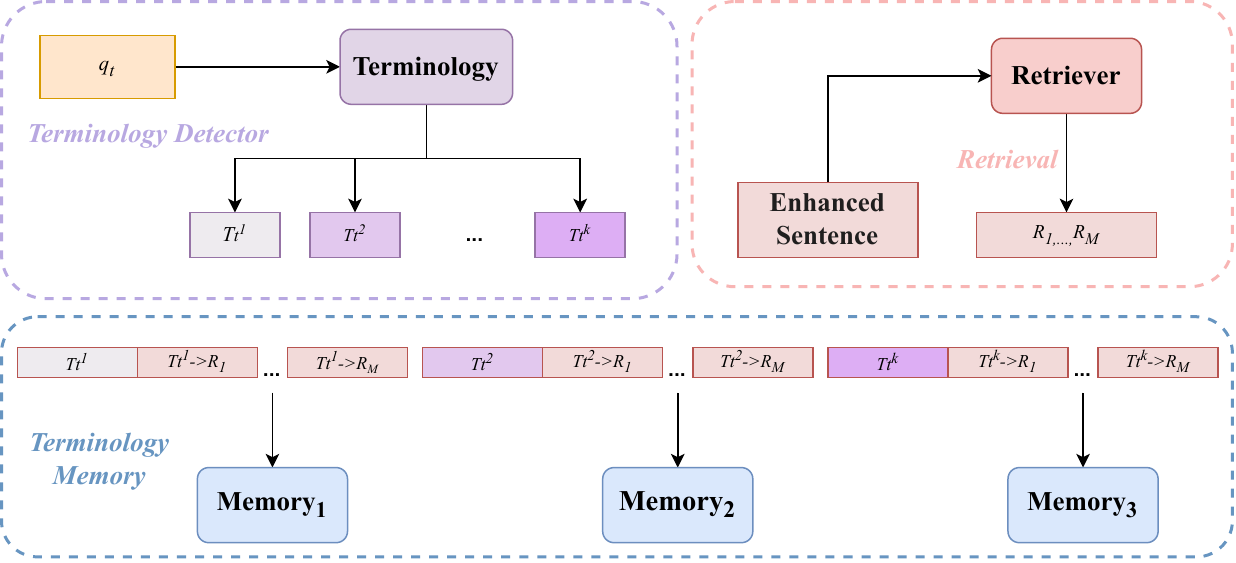}
% Alternatively, you can set a specific width or height.
% \includegraphics[width=1.5\linewidth]{QA table.pdf}
% \includegraphics[height=0.5\textheight]{QA table.pdf}
\caption{The overall pipeline of Terminology-Enhanced Information Retrieval (TEIR).}
\label{fig:TEIR}
\vspace{-4mm}
\end{figure}

\vspace{-4mm}
Our framework introduces an innovative approach to enhance the reliability of the generated content: (1) \textbf{Terminology-Enhanced Information Retrieval (TEIR)} constitutes the first part of our framework, focusing on the extraction and utilization of contextually significant information. The overall pipeline is shown in \ref{fig1}. 
(2) \textbf{Emotional In-Context Learning (EICL)} is the second part of our methodology, dedicated to understanding and integrating the emotional dimensions of the text. It leverages the generative capabilities of the framework, enriched with a nuanced understanding of emotional attributes.

% Both modules are intricately designed to cooperate, ensuring the generated text is contextually relevant, semantically rich, and emotionally resonant with the target attributes.

% \begin{figure}[htbp]
% \centering
% \includegraphics[scale=0.4]{Coordinate figure.pdf}
% \caption{Coordinate figure.}
% \label{fig:Coordinate figure}
% \end{figure}

\vspace{-2mm}
\subsection{Terminology-Enhanced Information Retrieval 
 (TEIR)}
\vspace{-2mm}

The first part of our framework, Terminology-Enhanced Information Retrieval, focuses on extracting and utilizing contextually significant information through a synergy of three core components in the enhancement:

\textbf{Terminology Detector (TD):} Utilize a Universal Information Extraction (UIE)-based model \(D\) to realize terminology detection, ensuring the retrieval process is sharply focused and aligned with the specified attributes.

\textbf{Terminology Memory (TM):}  After TD process we can get $N$ different terminologies $T = \{T_n\}^{N}_{n=1}$. However, in the actual medical knowledge base, different terminologies may correspond to the same meaning, so we have to do a simple filtering and alignment. We save the processed terms to the TM module, which ensures that subsequent questions can also be indexed to relevant terminologies and in turn allows the model to be progressively tuned as the dialogue progresses. 

It is worth noting that in this process we link the terminology in the TM one-to-one with the documents in the knowledge database, which will be used to speed up the subsequent retrieval process.

\textbf{Enhanced Sentence Generation (ESG):} 
A more accurate question specially generated for information retrieval can be achieved through terminology retrieval and combination using TM, reducing the effect of domain-specific vocabulary and facilitating robust information retrieval.

 After that, we perform Information Retrieval based on Enhanced Sentence. First, we link documents based on terminology $T = \{T_n\}^{k}_{n=1}$ in Enhanced Sentence, and the linked document is represented by $R = \{R_m\}^{M}_{m=1}$. After that, only the document in the link needs to be retrieved. 

 Considering the specialized nature of medical databases and the similarity of some documents, here we additionally use the following mapping to amplify the gap between the confidence probabilities for better document retriever.

\begin{equation}
\quad \hat{p}_n = map(p_n) =  \frac{\exp(p_n)}{\sum^N_{i=1} \exp(p_i)}, n=1,...,N
\end{equation}

% \vspace{-2mm}
\subsection{Emotional In-Context Learning}
\vspace{-2mm}
\begin{figure}[htbp]
\centering
\includegraphics[scale=0.4]{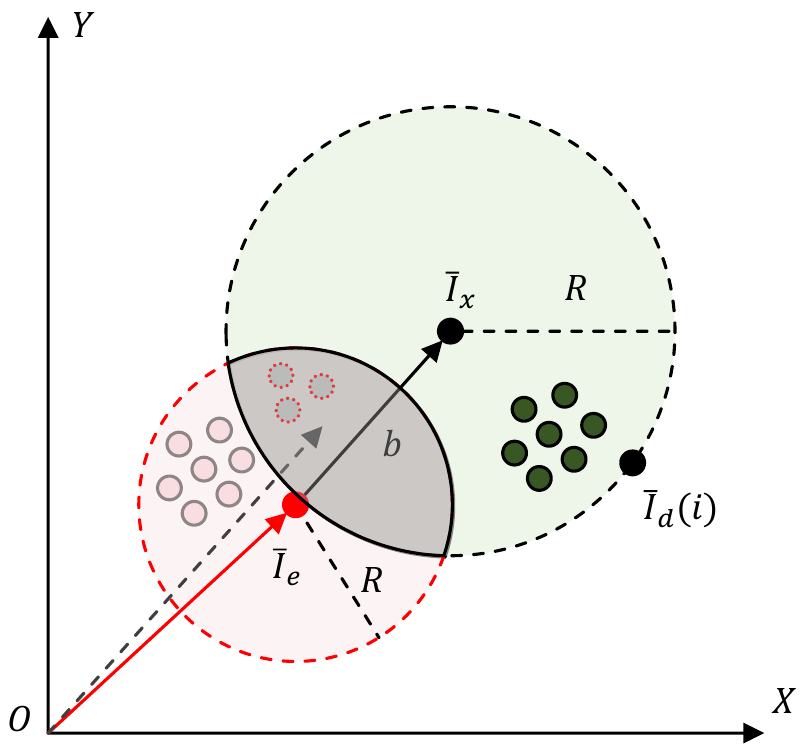}
\caption{The geometric diagram illustrates the emotion-influenced iterative generation process. Representations for the event schema, the document, and the \(I\)-th discrete demonstration are symbolized by \(\mathbf{I}_e\), \(\mathbf{I}_x\), and \(\mathbf{I}_d^{(i)}\) correspondingly.}
\vspace{-6mm}
\label{fig:Coordinate figure}
\end{figure}

Emotional In-Context Learning, the second part of our methodology, is dedicated to the integration and understanding of emotional dimensions:

\begin{enumerate}
    \item \textbf{Emotionally Attuned Generation:} The generative component \(G\) is tasked with sentence formation while ensuring that the text resonates with the desired emotional undertones, aligning with the target attributes.
    \item \textbf{Gradient-Guided Refinement:} Employs a gradient-guided retrieval-generation mechanism to adjust the generated text at each step, ensuring continual alignment with the target emotional attributes for contextually and emotionally coherent output.
\end{enumerate}

% \noindent
As shown in Figure~\ref{fig:Coordinate figure}, we can also understand the EICL in terms of a probability space. The regions \( I_e \), \( I_z \), and \( I_d^{(i)} \) are interpreted as states within the probability space. \( I_e \) represents the initial emotional context, \( I_z \) the influence of the document context, and \( I_d^{(i)} \) the iterative demonstration towards the true value. The overlapping areas and the vector \( R \) symbolize the adaptive process of the learning model.In each iteration, the system enhances the fidelity of learning outcome \( I_d^{(i)} \), bringing it nearer to the true value.Using such an effective guide, we can make the final generated content closer to the patient's space, which refers to a concept or psychological space that summarizes the patient's emotional and health background. It plays a significant role in predicting and aligning with the emotional and informational needs of patients in the model. To realize this, we adopt a suffix-tuning method to transform continuous tasks into discrete data usable for training while keeping the LM (Language Model) parameters unchanged. Mathematically speaking, this is the search for the optimal parameters $\phi$ that maximize the log-likelihood function given a training set $\mathcal{D}$ and fixed model parameters $\theta$.

\begin{equation} \label{eq-p}
\max_{\phi} P(y | x; \theta; \phi) = \max_{\phi} \sum_{y_i} P(y_i | h_{>i}; \theta; \phi)
\end{equation}

In equation \ref{eq-p}, $h_{>i}$ represents the set of all hidden states after the $i$-th position. If simplified, it can be considered as a state of the LM. $h_{>i}$ is obtained through the model $M_{\phi}$, which represents the state of the LM given the model parameters $\phi$. It can also be understood by the following example.

% \noindent
\textbf{Train:} Patient: \textit{\textcolor{blue}{I have a fever today;}} Doctor: \textit{\textcolor{orange}{Drink more hot water.}} We predict patient feedback: \textit{\textcolor{purple}{Negative.}}

% \noindent
\textbf{Test:} \textit{\textcolor{blue}{I have a fever today;}}

This dialogue serves as a training instance for the model to understand and predict emotional responses based on conversational context. Here, the simplicity of the doctor's advice in response to a possibly serious symptom might not address the patient's concerns adequately, leading to negative feedback. Emotional In-Context Learning aims to use such instances to teach the model the subtleties of emotional undercurrents in conversations, helping it predict and, in real-world applications, provide responses that are not only factually correct but also emotionally attuned to the patient's needs.

More specifically, here we construct a regenerated method using ECIL that utilizes sentiment to achieve better quality response content generation. In the table \ref{tab-T} we show a real case study.

\begin{table}[t]
\centering %把表居中
\vspace{-2mm}
\caption{Case study of regenerated method} %标题
\label{tab-T}
\setlength{\tabcolsep}{3mm}
\begin{tabular}{ll} %两个l代表该表一共两列，内容全部靠左
\toprule %第一道横线
\textbf{Input:} &  \textit{\textcolor{blue}{I have a fever today;}} \\
\midrule %第二道横线
\textbf{Answer:} & \textit{\textcolor{orange}{Drink more hot water.}} \\
\midrule %第三道横线
\textbf{Sentiment:} &  Negative. \\
\midrule %第四道横线
\textbf{Input:} & \textit{\textcolor{blue}{I have a fever today. Please do not say, "Drink more hot water.";}} \\
\textbf{Answer:} & \textit{\textcolor{orange}{Drink adequate amounts of fluids. }} \\
& \textit{\textcolor{orange} {Water, juices, clear soups, or hot lemonade are all good choices.}} \\
& \textit{\textcolor{orange} {Avoid caffeine and alcohol; these ingredients increase fluid loss.}} \\
\midrule %第五道横线
\textbf{Sentiment:} &  Positive. \\
\bottomrule %第六道横线
\end{tabular}
\end{table}

Considering the specialized nature of medical databases, some of the responses are difficult to be well understood by the consultants, and our approach can achieve increased consultant satisfaction through such context-based self-supervised feedback.

\section{Experiment}
\vspace{-2mm}
In this section, we first introduce the dataset, experimental setup, and baseline methods, followed by an exposition of the evaluation metrics and overall results, which include 803,564 medical consultation data entries. To ensure the efficacy of our model evaluation, this study also employs two metrics: BLEU-5 and ROUGE-2/L for assessment.

\vspace{-2mm}
\subsection{Experiment setup}
\vspace{-2mm}
\subsubsection{Datasets}
In order to develop and enhance the model's performance, our dataset utilizes a powerful web crawler system to gather a large dataset containing 803,564 medical consultation records. This dataset spans 12 distinct medical departments and includes more than 10 varied medical Q\&A scenarios, it encompasses patient inquiries about symptoms and treatments, health consultations, medical advice, and general medical knowledge. Furthermore, the data has been meticulously processed to ensure the removal of any personal identifiers, upholding patient confidentiality and adhering to strict privacy standards. These data are beneficial for training the model. Utilizing \(30\%\) of the dataset for testing purposes and allocating the remaining \(70\%\) for training the model significantly enhances the model's reliability. This distribution of data ensures a comprehensive learning process while maintaining a robust framework for validation and assessment of the model's performance.

% 给的数据集
% 硬件环境，V100；cpu xxx；gpu xxx；内存 xxx；
\subsubsection*{Evaluation Metrics.}
Our results are presented using a range of sophisticated evaluation metrics to ensure a comprehensive analysis of the model's performance in various aspects of language understanding and generation.

\begin{itemize}
    \item \textbf{BLEU-5 (Bilingual Evaluation Understudy):} This metric is an extension of the standard BLEU score, which is widely used to evaluate the quality of text that has been machine-translated from one language to another. BLEU-5 specifically refers to using the average precision of up to 5 grams (sequences of five consecutive words) to assess translation quality, providing a more nuanced understanding of the model's linguistic accuracy and fluency.
    
    \item \textbf{ROUGE-2/L (Recall-Oriented Understudy for Gisting Evaluation):} ROUGE is primarily used to evaluate automatic summarization and machine translation. ROUGE-2 refers to the overlap of bigrams between the generated text and a set of reference texts, effectively measuring the quality of content reproduction. ROUGE-L, on the other hand, focuses on the longest common subsequence and is particularly adept at assessing the sentence-level structure similarity between the generated text and reference texts.
\end{itemize}

In employing these advanced metrics, we aim to thoroughly assess the model's ability to generate coherent, contextually relevant, and linguistically accurate content. By focusing on both the micro-structure of language (as captured by BLEU-5) and the effective conveyance of meaning and structure (as captured by ROUGE-2/L), we provide a holistic evaluation of the model's performance in critical tasks of language processing and generation.

\vspace{-2mm}
\subsubsection{Baselines.} For comprehensive evaluation and fair comparison, we introduce five baseline models that are prominent in the field of language generation and attribute-based modeling.\textbf{DoctorGLM}\cite{Xiong2023DoctorGLMFY}, a specialized model for medical inquiries; \textbf{ChatGPT-3.5}\cite{Yldz2023ComparingRP}, an iteration of the GPT series with 3.5 billion parameters; \textbf{ChatGLM-6B}\cite{yang2023customizing}, a 6 billion parameter model designed for conversational tasks; \textbf{Ziya-LLaMA-13B}\cite{Lu2023ZiyaVisualBL}, optimized for linguistic tasks with 13 billion parameters; and \textbf{HuatuoGPT}\cite{Zhang2023HuatuoGPTTT}, our proprietary model before fine-tuning. Additionally, we showcase our model \textbf{T5}, which is \textbf{HuatuoGPT} after extensive fine-tuning to enhance its performance on a wide array of tasks.

\subsubsection{Experimental settings.} Our experimental setup is built on a robust hardware environment to facilitate efficient and reliable model training and evaluation. Specifically, we utilize NVIDIA-V100-32GB GPUs, known for their powerful computation capabilities, particularly in handling large-scale machine learning tasks. We initialize our models with the pre-trained T5 model, available in the HuggingFace Transformers library. We consider two model sizes, base and large, containing respectively 220M and 770M parameters. %We fine-tune the models on each dataset independently using AdamW (Loshchilov and Hutter, 2019) and conducted experiments on the specified V100 GPUs. 
Due to GPU memory limitations, we use different batch sizes for different models: 8 for T5-large and 16 for T5-base.

\begin{table}[t]
\centering 
\caption{AMT study results for medical consultant} 
\setlength{\tabcolsep}{3mm}
\begin{tabular}{lcccc} 
\toprule
\textbf{Model}& \textbf{Patient satisfaction} & & & \\
\midrule 
DoctorGLM & 5.67\% (34/600) & & & \\
ChatGPT-3.5 & 7.17\% (43/600) & & & \\
ChatGLM-6B & 2.83\% (17/600) & & & \\
Ziya-LLaMA-13B & 2.17\% (13/600) & & & \\
HuatuoGPT & \textbf{12.50\% (75/600)} & & & \\
\midrule 
Ours(w/o TEIR) & \textbf{17.50\% (105/600)} & & & \\
Ours(w/o EICL) & \textbf{15.33\% (92/600)} & & & \\
\textbf{Ours} & \textbf{36.83\% (221/600)} & & & \\
\bottomrule 
\end{tabular}
\label{tab1}
\end{table}
 
\subsection{Experiment result}

Table 1 presents the results of a study measuring patient satisfaction across various models. The first column lists the names of the models evaluated, including DoctorGLM, ChatGPT-3.5, ChatGLM-6B, Ziya-LLaMA-13B, HuatuoGPT, and variations of \textbf{Ours}. Each model represents a different approach or version in the medical consultation simulation or relevant language model application fields. The second column, labeled "Patient satisfaction," displays the percentage of patient satisfaction along with the raw number of satisfied instances out of 600. The satisfaction rates are expressed as percentages and are derived from a possibly larger-scale study involving 600 individual assessments or interactions for each model.

To ensure the integrity of the Amazon Mechanical Turk (AMT) study assessing patient satisfaction with medical consultation responses, participants were provided with unequivocal instructions. Satisfaction was gauged using a 5-point Likert scale, ranging from 1 (\textit{not satisfied}) to 5 (\textit{extremely satisfied}). Quality control was maintained through embedded attention checks to validate participant engagement. Consistency in responses was assured by interspersing identical questions in varying forms throughout the survey, enabling cross-verification of data reliability. These methodologies combined to form a robust framework, optimizing the authenticity of the satisfaction metrics obtained.

Notably, the patient satisfaction percentages progressively increase for the variants of \textbf{Ours} - \textbf{Ours (w/o TEIR)}, \textbf{Ours (w/o EICL)}, and \textbf{Ours}, with \textbf{Ours} achieving the highest satisfaction rate at 36.83\% (221 out of 600). This is attributed to the positive impact of the TEIR and EICL features on the \textbf{Ours} model's ability to attract and satisfy patients. The inclusion and optimization of these features in the final iteration of \textbf{Ours} evidently contribute to its superior performance in patient satisfaction, underscoring the significance of continuous model enhancement and feature integration in the development of medical consultation models.

In the evaluation of various language models, as displayed in Table 2, we have witnessed significant advancements in model performance on the Chinese medical QA dataset. The table outlines the results for a series of models, including T5 (fine-tuned),the previously five model, and \textbf{Ours} model.  It's noteworthy that \textbf{Ours}, which is an enhanced iteration based on the fundamental approach of HuatuoGPT through the Integrated Argument Generation (IAG) enhancement, demonstrates marked improvements over its counterparts, particularly in metrics such as BLEU-1, GLEU, ROUGE-1, and Distinct-2—critical indicators of the model's linguistic precision and contextual understanding, as shown in Table 2. The results indicate a significant improvement of \textbf{Ours} over the state-of-the-art (SOTA) models (24.68 to 26.12 for BLEU-1 and 8.07 to 8.23 for GLEU), and a comparable leap over another SOTA improvement (27.93 to 29.07 for ROUGE-1 and 0.93 to 0.96 for Distinct-2).

Compared to ChatGLM variants, our model shows a significant advantage; however, DoctorGLM performs poorly on BLEU and GLEU metrics. This could be due to the model's potential difficulty in handling the subtleties of context within medical texts. Medical language is typically laden with jargon and can be quite nuanced. If DoctorGLM is not fine-tuned on medical-specific data, it may fail to capture the complexity required for higher BLEU and GLEU scores. Nonetheless, it performs well on the Distinct metrics. This is because DoctorGLM might employ a generation strategy that favors diversity over precision. Such an approach might manifest in a model that is less likely to repeat the same phrases and more inclined to explore the breadth of its training data when generating responses. In summary, our model outperforms nearly all evaluated metrics when compared to others.

\begin{figure}
\vspace{-4mm}
	\centering
	\subfigure[GPT-4 Evaluation] {\includegraphics[width=.49\textwidth]{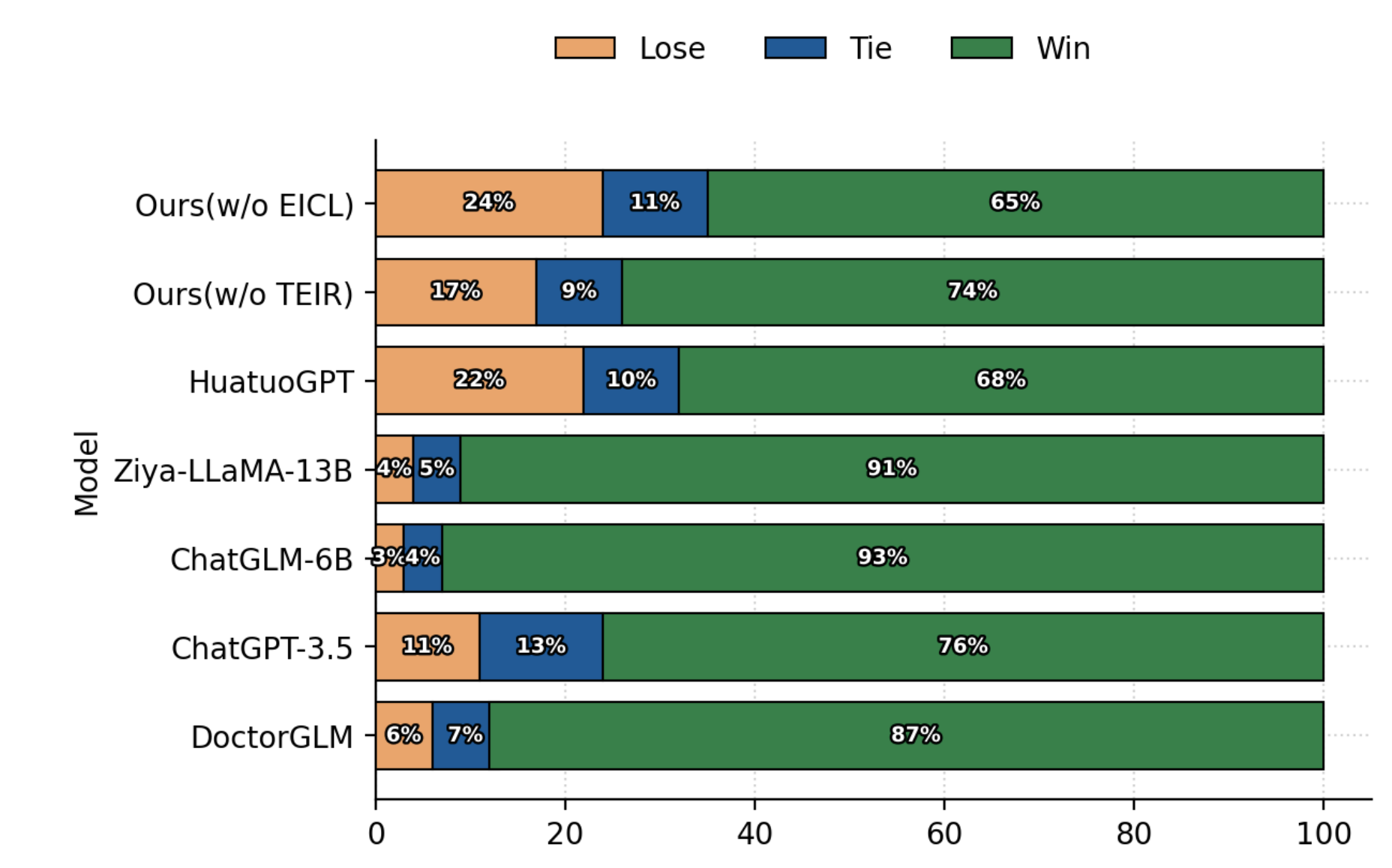}}
	\subfigure[Doctor Evaluation] {\includegraphics[width=.49\textwidth]{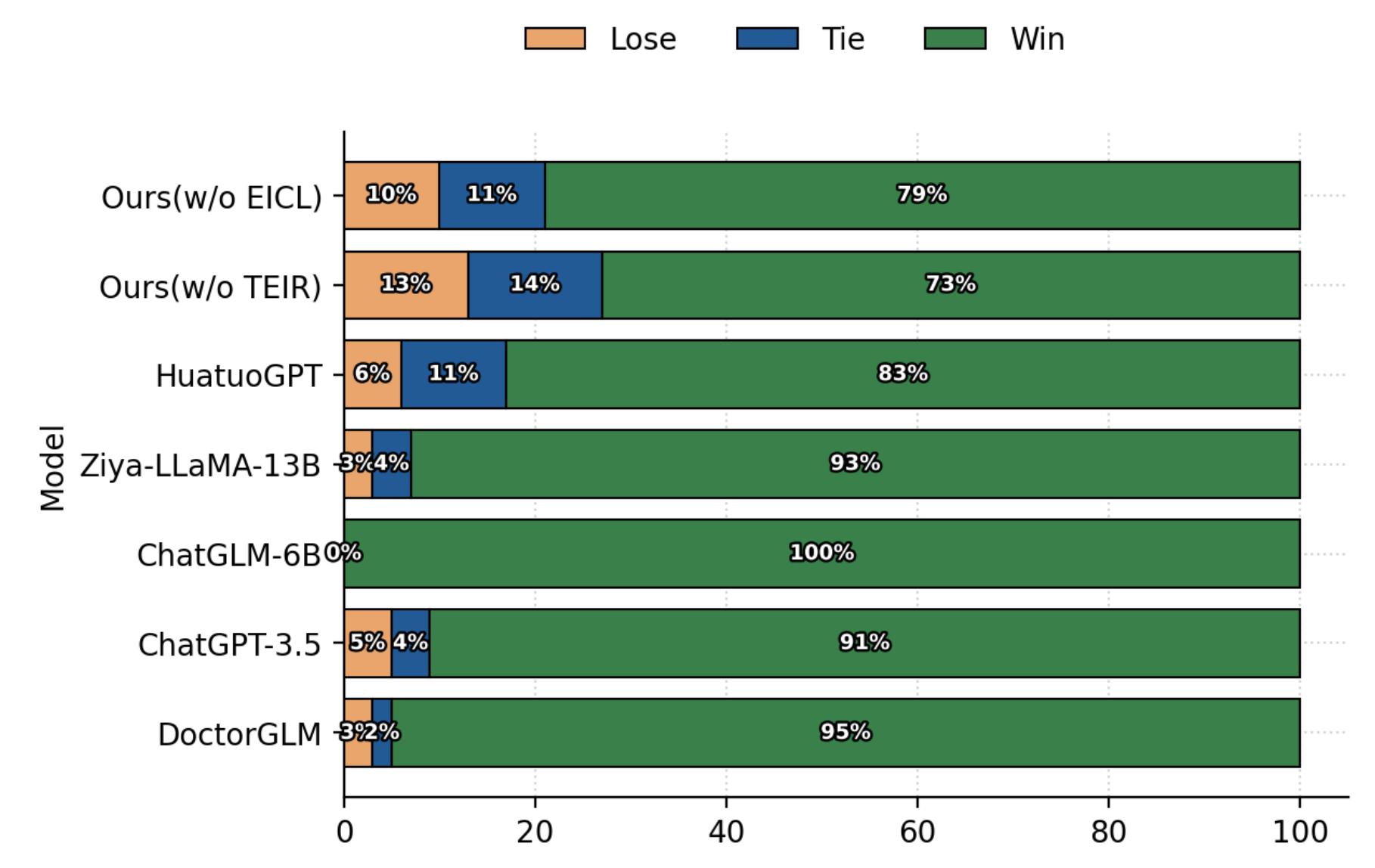}}
	\caption{1v1 battle between different models and ours}
	\label{fig_E1}
\vspace{-6mm}
\end{figure}

In order to make a more direct comparison of different models, we used GPT-4 and a professional doctor to evaluate our model and other models one-on-one. It can be seen that our model has a significant advantage over the others; doctorGLM does not perform well on some of the quantitative metrics, but when viewed by a professional doctor, it has a very good performance, probably because the content it generates is more in line with the human mind. Overall, our model won almost all 1v1 battles.
\vspace{-2mm}
\subsection{Ablation Study}
% \vspace{-8mm}
The ablation study focuses on two components of the \textbf{Ours} model: TEIR and EICL. By analyzing the performance impact of each component's removal, we gain insights into their contributions.

\vspace{-5mm}
\subsubsection{Ablation of TEIR}
Without TEIR, \textbf{Ours (w/o TEIR)}exhibits a decrease in BLEU-1 and BLEU-4 scores, indicating the importance of TEIR in capturing accurate unigrams and coherent longer n-gram sequences. The reduction in the GLEU score suggests that TEIR enhances sentence-level alignment with reference translations. Lower ROUGE scores point to TEIR's role in replicating essential content and maintaining structural integrity.
\vspace{-5mm}
\subsubsection{Ablation of EICL}
The absence of EICL in \textbf{Ours (w/o EICL)} leads to lowered performance across ROUGE and Distinct metrics. This suggests that EICL may be instrumental in capturing critical details and enhancing the diversity and uniqueness of the generated text.
\vspace{-5mm}
\subsubsection{Combined Impact}
The results of the ablation study indicate that both TEIR and EICL significantly contribute to the \textbf{Ours} model's ability to handle complex tasks in the Chinese medical QA dataset. Their combined contributions are pivotal in enhancing the model's precision and capacity to generate contextually rich and diverse language.

\begin{table}[t]
\centering 
\caption{Test Result on Chinese medical QA dataset} 
\label{tab2}
\setlength{\tabcolsep}{1mm}
\begin{tabular}{lccccc} 
\toprule 
\textbf{Model} & \textbf{BLEU-1} & \textbf{BLEU-2} & \textbf{BLEU-3} & \textbf{BLEU-4} & \textbf{GLEU} \\
\midrule 
T5 (fine-tuned) & 21.27 & 13.65 & 9.89 & 7.21 & 9.07 \\
DoctorGLM & 9.89 & 5.36 & 2.56 & 1.45 & 4.43 \\
ChatGPT-3.5 & 17.87 & 6.54 & 2.37 & 0.99 & 4.87 \\
ChatGLM-6B & \underline{23.53} & \underline{12.07} & \underline{6.54} & 3.65 & \textbf{8.07} \\
Ziya-LLaMA-13B & 22.01 & 11.34 & 6.35 & \underline{4.01} & 6.99 \\
HuatuoGPT & \textbf{24.68} & \textbf{13.25} & \textbf{7.88} & \textbf{4.34} & \underline{7.30} \\
Ours(w/o TEIR) & 25.54 & 13.45 & 8.01 & 4.64 & 7.56 \\
Ours(w/o EICL) & \underline{25.73} & \underline{14.01} & \underline{8.16} & \underline{4.97} & \underline{8.12} \\
\textbf{Ours} & \textbf{26.12} & \textbf{14.34} & \textbf{8.43} & \textbf{5.01} & \textbf{8.23} \\
\midrule 
\textbf{Model} & \textbf{ROUGE-1} & \textbf{ROUGE-2} & \textbf{ROUGE-L} & \textbf{Distinct-1} & \textbf{Distinct-2} \\
\midrule 
T5 (fine-tuned) & 30.08 & 13.76 & 24.43 & 0.39 & 0.50 \\
DoctorGLM & 22.79 & 5.43 & 11.73 & \textbf{0.82} & \textbf{0.93} \\
ChatGPT-3.5 & 19.79 & 2.46 & 12.76 & 0.67 & 0.89 \\
ChatGLM-6B & \underline{27.69} & 6.17 & \textbf{17.98} & 0.65 & 0.87 \\
Ziya-LLaMA-13B & 27.41 & \underline{6.90} & 13.24 & 0.75 & \underline{0.91} \\
HuatuoGPT & \textbf{27.93} & \textbf{7.28} & \underline{15.20} & \underline{0.74} & \textbf{0.93} \\
Ours(w/o TEIR) & \underline{29.13} & 8.10 & 14.01 & \underline{0.81} & \underline{0.93} \\
Ours(w/o EICL) & 28.23 & \underline{7.69} & \underline{13.32} & 0.76 & 0.90 \\
\textbf{Ours} & \textbf{29.07} & \textbf{8.10} & \textbf{14.50} & \textbf{0.84} & \textbf{0.96} \\
\bottomrule 
\end{tabular}
\end{table}
\vspace{-2mm}

% 生成2000条；测试了BLEU指标
% 最好的指标加粗，textbf；第二好，下划线；
% Ours(w/o TEIR)；Ours(w/o EICL)；
\vspace{-2mm}
\section{Conclusion}
\vspace{-2mm}
In conclusion, our method is a retrieval-augmented generation framework. It innovatively integrates Terminology-Enhanced Information Retrieval and Emotional In-Context Learning, significantly enhancing its ability to understand complex medical dialogues and emotional contexts. Following rigorous evaluation with 803,564 medical consultations, our method has demonstrated superior performance compared to existing models. This advancement represents a substantial contribution to the field of artificial intelligence, particularly in retrieval-augmented methodologies. It holds immense potential to revolutionize medical information retrieval and generation, setting a new trajectory for the application of artificial intelligence in healthcare. This research highlights the transformative implications of our method, not just as a technological innovation but as a pivotal step towards more sophisticated, responsive, and empathetic healthcare solutions. Given the limited time doctors spend with each patient and ongoing patient complaints, satisfaction with medical consultations is not consistently achieved. Our new framework has the potential to create a more satisfying medical consultation environment for patients by addressing their emotions. Additionally, our framework promises to contribute to future medical education, offering insights on establishing effective patient relationships, which could lead to higher patient satisfaction rates.
%
% ---- Bibliography ----
%
% BibTeX users should specify bibliography style 'splncs04'.
% References will then be sorted and formatted in the correct style.
%
\bibliographystyle{unsrt}
% \bibliography{mybibliography}
%
\clearpage
\bibliography{sample}

\begin{thebibliography}{10}

\bibitem{schwartz2023black}
Ilan~S Schwartz, Katherine~E Link, Roxana Daneshjou, and Nicol{\'a}s Cort{\'e}s-Penfield.
\newblock Black box warning: large language models and the future of infectious diseases consultation.
\newblock {\em Clinical Infectious Diseases}, page ciad633, 2023.

\bibitem{wang2023time}
Yuxin Wang and Shunda Du.
\newblock Time to rebuild the doctor-patient relationship in china.
\newblock {\em Hepatobiliary Surgery and Nutrition}, 12(2):235, 2023.

\bibitem{younis2024systematic}
Hussain~A Younis, Taiseer Abdalla~Elfadil Eisa, Maged Nasser, Thaeer~Mueen Sahib, Ameen~A Noor, Osamah~Mohammed Alyasiri, Sani Salisu, Israa~M Hayder, and Hameed~AbdulKareem Younis.
\newblock A systematic review and meta-analysis of artificial intelligence tools in medicine and healthcare: Applications, considerations, limitations, motivation and challenges.
\newblock {\em Diagnostics}, 14(1):109, 2024.

\bibitem{engelseth2021systems}
Per Engelseth, BE~White, Ingunn Mundal, Trude~Fl{\o}ystad Eines, and Duangpun Kritchanchai.
\newblock Systems modelling to support the complex nature of healthcare services.
\newblock {\em Health and Technology}, 11:193--209, 2021.

\bibitem{alberts2023large}
Ian~L Alberts, Lorenzo Mercolli, Thomas Pyka, George Prenosil, Kuangyu Shi, Axel Rominger, and Ali Afshar-Oromieh.
\newblock Large language models (llm) and chatgpt: what will the impact on nuclear medicine be?
\newblock {\em European journal of nuclear medicine and molecular imaging}, 50(6):1549--1552, 2023.

\bibitem{el2023integration}
Abdulmotaleb El~Saddik and Sara Ghaboura.
\newblock The integration of chatgpt with the metaverse for medical consultations.
\newblock {\em IEEE Consumer Electronics Magazine}, 2023.

\bibitem{senbekov2020recent}
Maksut Senbekov, Timur Saliev, Zhanar Bukeyeva, Aigul Almabayeva, Marina Zhanaliyeva, Nazym Aitenova, Yerzhan Toishibekov, Ildar Fakhradiyev, et~al.
\newblock The recent progress and applications of digital technologies in healthcare: a review.
\newblock {\em International journal of telemedicine and applications}, 2020, 2020.

\bibitem{wang2023ethical}
Changyu Wang, Siru Liu, Hao Yang, Jiulin Guo, Yuxuan Wu, and Jialin Liu.
\newblock Ethical considerations of using chatgpt in health care.
\newblock {\em Journal of Medical Internet Research}, 25:e48009, 2023.

\bibitem{kenei2022supporting}
Jonah Kenei.
\newblock {\em Supporting Information Retrieval From Clinical Narrative Texts Using Text Classification and Visualization Techniques}.
\newblock PhD thesis, University of Nairobi, 2022.

\bibitem{ho2022explainability}
Tin~Kam Ho, Yen-Fu Luo, and Rodrigo~Capobianco Guido.
\newblock Explainability of methods for critical information extraction from clinical documents: A survey of representative works.
\newblock {\em IEEE Signal Processing Magazine}, 39(4):96--106, 2022.

\bibitem{van2020online}
Piet van~der Keylen, Johanna Tomandl, Katharina Wollmann, Ralph Moehler, Mario Sofroniou, Andy Maun, Sebastian Voigt-Radloff, and Luca Frank.
\newblock The online health information needs of family physicians: systematic review of qualitative and quantitative studies.
\newblock {\em Journal of medical Internet research}, 22(12):e18816, 2020.

\bibitem{Ren2023RetrieveandSampleDE}
Yubing Ren, Yanan Cao, Ping Guo, Fang Fang, Wei Ma, and Zheng Lin.
\newblock Retrieve-and-sample: Document-level event argument extraction via hybrid retrieval augmentation.
\newblock In {\em Annual Meeting of the Association for Computational Linguistics}, 2023.

\bibitem{jiang2023think}
Xinke Jiang, Ruizhe Zhang, Yongxin Xu, Rihong Qiu, Yue Fang, Zhiyuan Wang, Jinyi Tang, Hongxin Ding, Xu~Chu, Junfeng Zhao, and Yasha Wang.
\newblock Think and retrieval: A hypothesis knowledge graph enhanced medical large language models, 2023.

\bibitem{li2023chatdoctor}
Yunxiang Li, Zihan Li, Kai Zhang, Ruilong Dan, Steve Jiang, and You Zhang.
\newblock Chatdoctor: A medical chat model fine-tuned on a large language model meta-ai (llama) using medical domain knowledge, 2023.

\bibitem{yu2023leveraging}
Ping Yu, Hua Xu, Xia Hu, and Chao Deng.
\newblock Leveraging generative ai and large language models: A comprehensive roadmap for healthcare integration.
\newblock In {\em Healthcare}, volume~11, page 2776. MDPI, 2023.

\bibitem{ford2016extracting}
Elizabeth Ford, John~A Carroll, Helen~E Smith, Donia Scott, and Jackie~A Cassell.
\newblock Extracting information from the text of electronic medical records to improve case detection: a systematic review.
\newblock {\em Journal of the American Medical Informatics Association}, 23(5):1007--1015, 2016.

\bibitem{malet1999model}
Gary Malet, Felix Munoz, Richard Appleyard, and William Hersh.
\newblock A model for enhancing internet medical document retrieval with “medical core metadata”.
\newblock {\em Journal of the American Medical Informatics Association}, 6(2):163--172, 1999.

\bibitem{zhang2023iag}
Zhebin Zhang, Xinyu Zhang, Yuanhang Ren, Saijiang Shi, Meng Han, Yongkang Wu, Ruofei Lai, and Zhao Cao.
\newblock Iag: Induction-augmented generation framework for answering reasoning questions.
\newblock In {\em Proceedings of the 2023 Conference on Empirical Methods in Natural Language Processing}, pages 1--14, 2023.

\bibitem{varshney2023knowledge}
Deeksha Varshney, Aizan Zafar, Niranshu~Kumar Behera, and Asif Ekbal.
\newblock Knowledge grounded medical dialogue generation using augmented graphs.
\newblock {\em Scientific Reports}, 13(1):3310, 2023.

\bibitem{vanAtteveldt2021TheVO}
Wouter van Atteveldt, Mariken~A.C.G. van~der Velden, and Mark Boukes.
\newblock The validity of sentiment analysis: Comparing manual annotation, crowd-coding, dictionary approaches, and machine learning algorithms.
\newblock {\em Communication Methods and Measures}, 15:121 -- 140, 2021.

\bibitem{Yang2021ReExaminingTI}
Ming Yang, Jinglu Jiang, Melody~Y. Kiang, and Fangyun Yuan.
\newblock Re-examining the impact of multidimensional trust on patients’ online medical consultation service continuance decision.
\newblock {\em Information Systems Frontiers}, 24:983 -- 1007, 2021.

\bibitem{Allaith2021AraSenCorpusAS}
Ali Al-laith, Muhammad Shahbaz, Hind Alaskar, and Asim Rehmat.
\newblock Arasencorpus: A semi-supervised approach for sentiment annotation of a large arabic text corpus.
\newblock {\em Applied Sciences}, 2021.

\bibitem{inproceedings}
Zhihua Wen, Zhiliang Tian, Zhen Huang, Yuxin Yang, Zexin Jian, Changjian Wang, and Dongsheng li.
\newblock Grace: Gradient-guided controllable retrieval for augmenting attribute-based text generation.
\newblock pages 8377--8398, 01 2023.

\bibitem{Wan2021InfluencingFA}
Yan Wan, Ziqing Peng, Yalu Wang, Yifan Zhang, Jinping Gao, and Baojun Ma.
\newblock Influencing factors and mechanism of doctor consultation volume on online medical consultation platforms based on physician review analysis.
\newblock {\em Internet Res.}, 31:2055--2075, 2021.

\bibitem{Gidwani2017ImpactOS}
Risha Gidwani, Cathina~T. Nguyen, Alexis Kofoed, Catherine Carragee, Tracy~A Rydel, Ian Nelligan, Amelia~L Sattler, Megan Mahoney, and Steven Lin.
\newblock Impact of scribes on physician satisfaction, patient satisfaction, and charting efficiency: A randomized controlled trial.
\newblock {\em The Annals of Family Medicine}, 15:427 -- 433, 2017.

\bibitem{Jiang2020AnalysisOM}
Jinglu Jiang, Ann~Frances Cameron, and Ming Yang.
\newblock Analysis of massive online medical consultation service data to understand physicians’ economic return: Observational data mining study.
\newblock {\em JMIR Medical Informatics}, 8, 2020.

\bibitem{Xiong2023DoctorGLMFY}
Honglin Xiong, Sheng Wang, Yitao Zhu, Zihao Zhao, Yuxiao Liu, Linlin Huang, Qian Wang, and Dinggang Shen.
\newblock Doctorglm: Fine-tuning your chinese doctor is not a herculean task.
\newblock {\em ArXiv}, abs/2304.01097, 2023.

\bibitem{Yldz2023ComparingRP}
Mustafa~Said Yıldız.
\newblock Comparing response performances of chatgpt-3.5, chatgpt-4 and bard to health-related questions: Comprehensiveness, accuracy and being up-to-date.
\newblock {\em SSRN Electronic Journal}, 2023.

\bibitem{yang2023customizing}
Bang Yang, Asif Raza, Yuexian Zou, and Tong Zhang.
\newblock Customizing general-purpose foundation models for medical report generation, 2023.

\bibitem{Lu2023ZiyaVisualBL}
Junyu Lu, Di~Zhang, Xiaojun Wu, Xinyu Gao, Ruyi Gan, Jiaxing Zhang, Yan Song, and Pingjian Zhang.
\newblock Ziya-visual: Bilingual large vision-language model via multi-task instruction tuning.
\newblock {\em ArXiv}, abs/2310.08166, 2023.

\bibitem{Zhang2023HuatuoGPTTT}
Hongbo Zhang, Junying Chen, Feng Jiang, Fei Yu, Zhihong Chen, Jianquan Li, Guimin Chen, Xiangbo Wu, Zhiyi Zhang, Qingying Xiao, Xiang Wan, Benyou Wang, and Haizhou Li.
\newblock Huatuogpt, towards taming language model to be a doctor.
\newblock In {\em Conference on Empirical Methods in Natural Language Processing}, 2023.

\end{thebibliography}

\end{document}